# Accountable Human-AI Deliberation with LLMs: Scaling Collective Intelligence through Symbiotic Scaffolding

**Wajdi Zaghouani**
Northwestern University in Qatar
wajdi.zaghouani@northwestern.edu

**Abstract**

Large language models (LLMs) can support democratic deliberation at scales previously constrained by turn-taking and facilitation bandwidth. Recent work shows that LLM-generated group statements are often preferred over human-mediated outputs, while theoretical analyses argue that LLMs relax the simultaneity constraints limiting collective intelligence. Yet pure LLM mediation risks collapsing pluralism, over-optimizing for agreement, and undermining legitimacy when participants cannot contest how they are represented. We propose a symbiotic human-AI framework organized into three layers: observation and diversity amplification, facilitation with clause-level provenance, and human primacy for ratification. Our contributions include graded coverage, diversity, and erasure metrics with salience-aware weighting; a provenance pipeline combining cross-encoder similarity with causal knockout diagnostics; preference-conditioned trade-off control; equity-aware contestability workflows; adversarial robustness tests; and an evaluation protocol with ablation designs informed by evidence of LLM-as-judge limitations. The result is a testable blueprint for deliberation technology that scales collective intelligence while preserving agency and legitimacy.

## 1. Introduction

Deliberation is a communicative process through which groups exchange reasons, weigh arguments, and seek decisions that can be justified to those bound by them. The normative ideal transcends mere aggregation of preferences, aiming instead for mutual justification and learning under conditions of inclusion and respect (Habermas, 1984). Yet scaling deliberation beyond small groups has long posed formidable challenges. Even with digital platforms, typical formats inherit sequential turn-taking, facilitator scarcity, and predictable distortions such as dominance by high-status speakers, early consensus pressure, and information overload (Fishkin, 2011).

Large language models fundamentally change these engineering boundary conditions. The Habermas Machine operationalizes a mediator role in which participants provide opinions and critiques, the system generates candidate group statements, and a learned reward model selects revisions likely to maximize endorsement (Tessler et al., 2024). In evaluations with over 5,700 participants, AI-generated statements were frequently preferred to human-mediated statements, and groups showed reduced polarization after AI-mediated deliberation. Page argues that LLMs alter the “physics” of collective intelligence by enabling parallel input processing at unprecedented scale (Page, 2025). Where traditional deliberation required one speaker at a time, LLM-based systems can ingest many inputs in parallel and return structured representations. This reframes the design problem from “how do we let everyone talk?” to “how do we preserve diversity, contestability, and meaning when everyone can contribute at once?”

However, recent critiques reveal that optimizing for agreement can narrow deliberation into a product rather than a legitimate process (Volpe, 2025; Palomo Hernández, 2025; Oleart and Palomo Hernández, 2025). Risks include shallow consensus that ignores value conflicts, loss of emotional and narrative dimensions, and opacity regarding whose views shaped which phrases. Furthermore, benchmarking work on multi-LLM deliberation protocols demonstrates that complex aggregation schemes may dramatically underperform simpler best-of-N selection approaches (Kaushal and Singh, 2025), cautioning against the assumption that complexity improves quality.

This paper proposes a *symbiotic* framework that addresses both promise and pitfalls. Symbiosis denotes explicit division of labor with accountability: AI provides scale, structure, and reflective mirrors of group discourse, while humans retain primacy for interpretation, contestation, and final endorsement. The framework is presented as a testable blueprint with sufficient technical specification to enable implementation and controlled evaluation. We make the following contributions:

1. Formal definitions of coverage, diversity, and erasure metrics with graded scoring and salience-aware minority weighting (Section 4.2.1).
2. A clause-level provenance pipeline combining cross-encoder similarity with causal diagnostics including knockout regeneration (Section 4.2.3).
3. Discussion of how preference-conditioned alignment mechanisms such as PARM (Lin et al.,

2025) enable controllable trade-offs at inference time (Section 4.2.2).
4. Concrete contestability workflows with equity-aware rate limits and governance protocols, illustrated with a worked example (Section 4.3.1).
5. Robustness tests including adversarial attribution attacks, informed by recent work on adversarially robust authorship segmentation (Sai Teja et al., 2025) (Section 4.4).
6. An evaluation protocol with ablation designs and validated psychometric instruments, informed by LLM-as-judge limitations (Li et al., 2025) (Section 6).

## 2. Background and Related Work

### 2.1. Deliberation and Collective Intelligence

Habermas frames communicative action as coordination through language oriented toward mutual understanding rather than strategic manipulation (Habermas, 1984). His discourse ethics holds that valid norms are those to which all affected parties could agree as participants in rational discourse. Fishkin's deliberative polling engages representative samples in structured deliberation, demonstrating that approximately 70% of participants change their views when they believe their voice matters (Fishkin, 2018; Fishkin et al., 2024). Collective intelligence research shows that groups exhibit a measurable factor predicting performance across cognitive domains, emerging from interaction patterns including social sensitivity and equality of turn-taking (Woolley et al., 2010).

Riedl and De Cremer argue that AI can support collective intelligence by augmenting collective memory, attention, and reasoning (Riedl and De Cremer, 2025). However, they caution that AI deployment carries risks of deskilling, homogenization, bias amplification, and reduction of intellectual diversity. Network science further illuminates how communication structure shapes whether groups explore diverse solutions or converge prematurely (Centola, 2022): dense networks accelerate convergence but may suppress alternatives, while sparser networks preserve diversity but slow coordination. These findings motivate a key design principle: improving collective intelligence requires not only better summaries but preserving conditions under which diverse ideas can surface and be integrated without coercion.

### 2.2. LLM Mediation: Results and Critiques

The Habermas Machine employs two LLMs: a generative model producing candidate statements and a reward model predicting endorsement (Tessler et al., 2024). Through iterative refinement incorporating participant opinions and critiques, the system searches for statements likely to achieve broad agreement. More than 56% of participants preferred AI statements, and the system incorporated dissenting voices rather than simply appealing to majorities. However, Volpe argues that consensus optimization may fundamentally conflict with deliberation as communicative autonomy (Volpe, 2025). Drawing on Habermas's own discourse theory, Volpe notes that Habermas does not argue deliberation should be structured toward the telos of consensus; rather, consensus serves as a precondition for the kind of discourse that should guide deliberation.

Palomo Hernández examines the normative choices embedded in system design, arguing that optimization targets, training data, and aggregation methods all encode particular conceptions of what deliberation should accomplish (Palomo Hernández, 2025). Oleart and Palomo Hernández situate these concerns within a political economy critique, arguing that technosolutionism reinforces depoliticization and disintermediation (Oleart and Palomo Hernández, 2025). A recurring theme across these critiques is that endorsement alone is an insufficient measure of deliberative quality; participants must also be able to trace, contest, and reshape how their contributions are represented.

### 2.3. Multi-Objective Alignment at Test Time

A critical implementation pathway for multi-objective selection involves preference-conditioned alignment. Lin et al. propose PARM, a unified reward model that conditions on preference vectors to control trade-offs during inference without retraining (Lin et al., 2025). Its PBLoRA mechanism modulates outputs based on user-specified preference weights, achieving strong alignment at lower cost than training separate reward models. This capability is central to the trade-off control we propose in Section 4.2.2.

### 2.4. Multi-Agent Deliberation: Cautionary Evidence

DeliberationBench evaluates multi-LLM deliberation protocols against a best-of-N selection baseline (Kaushal and Singh, 2025). The baseline outperforms the best deliberation protocol by a factor of six ($p < 0.01$) at lower computational cost. We do not propose autonomous multi-LLM deliberation but rather LLM-assisted observation and synthesis with human oversight. This finding motivates parsimony in our aggregation design.

### 2.5. LLM-as-Judge Limitations

CounselBench demonstrates through 2,000 expert evaluations that LLM judges consistently overrate model responses, overlook safety issues, and diverge sharply from human preferences at the span level (Li et al., 2025). This directly informs our evaluation protocol: we recommend human raters with span-level audit capabilities for assessing representation quality (Section 6).

### 2.6. Human-AI Deliberation, Argument Mining, and Provenance

Ma et al. introduce a Human-AI Deliberation framework where an LLM serves as a communication bridge for iterative decision updates, demonstrating that structured disagreement improves task performance over conventional explainable AI (Ma et al., 2025). In computational argumentation, methods for extracting claims, premises, and attack/support relations offer richer representations than flat clustering (Chen et al., 2024). Work on Arabic competitive debate corpora has shown that argument structure extraction can be productively applied to non-English deliberative discourse, demonstrating the cross-linguistic applicability of these techniques (Khader et al., 2024). Platforms like Polis enable large-scale opinion mapping (Small et al., 2023). Our framework extends this line of work with formal provenance tracking combining similarity and causal diagnostics, preference-conditioned multi-objective control, and explicit contestability mechanisms. Recent work on adversarially robust authorship segmentation (Sai Teja et al., 2025) informs our provenance robustness layer, and the Guaranteed Safe AI framework (Dalrymple et al., 2024) motivates our adoption of verifiable safety properties checkable over system logs.

## 3. Problem Formulation and Requirements

We model a deliberation episode as a sequence of rounds. In each round $t$, participants submit contributions $x^t$ that may include opinions, reasons, evidence, narratives, or critiques. The system maintains a shared representation $R^t$ comprising: a topic and stance map with clusters $C^t = \{C^t_1, \ldots, C^t_K\}$, candidate synthesis statements $S^t$, and a provenance structure $P^t$ linking statement clauses to contributing inputs.

We define eight design requirements motivated by empirical demonstrations and legitimacy critiques. Each requirement is justified below with reference to the specific evidence or critique that motivates it.

**Scalability (R1)** supports hundreds to thousands of participants with bounded facilitator load. *Justification:* Traditional deliberation formats face hard ceilings on participation; Fishkin's deliberative polling typically involves 200 to 500 participants, beyond which facilitation becomes impractical (Fishkin, 2018). Page demonstrates that LLMs can relax simultaneity constraints, but only if the system is designed for parallel input (Page, 2025).

**Diversity Retention (R2)** preserves minority perspectives as first-class elements rather than footnotes. *Justification:* Riedl and De Cremer identify homogenization as a core risk of AI-augmented collective intelligence (Riedl and De Cremer, 2025). Centola shows that dense communication networks suppress alternatives (Centola, 2022). Deliberation systems must actively counteract these tendencies.

**Contestability (R3)** enables participants to challenge representation and request alternative framings. *Justification:* Volpe argues that deliberation requires communicative autonomy, which is violated when participants cannot contest how their views are represented (Volpe, 2025). Palomo Hernández shows that design choices in automated systems encode normative assumptions that must be challengeable (Palomo Hernández, 2025).

**Provenance (R4)** exposes traceable links from phrasing to inputs, combining correlation-based and causal evidence. *Justification:* Oleart and Palomo Hernández argue that opacity in automated deliberation systems is a form of depoliticization (Oleart and Palomo Hernández, 2025). CounselBench demonstrates that LLM outputs can diverge from source material in ways that humans detect but automated evaluators miss (Li et al., 2025).

**Reframing Support (R5)** surfaces value conflicts beyond surface compromise. *Justification:* Genuine deliberation often involves reconceptualizing problems rather than splitting differences. Ma et al. show that structured disagreement improves task performance, suggesting that surfacing conflict is productive rather than destructive (Ma et al., 2025).

**Human Primacy (R6)** requires explicit human ratification with veto rights. *Justification:* Oleart and Palomo Hernández argue that technosolutionism replaces political agency with optimized text (Oleart and Palomo Hernández, 2025). The framework must ensure that AI remains facilitative, never decisional.

**Robustness (R7)** addresses manipulation, prompt injection, and adversarial attribution gaming. *Justification:* DAMASHA demonstrates that adversarial attacks can defeat text attribution systems (Sai Teja et al., 2025). DeliberationBench shows that multi-agent protocols are vulnerable to cascading errors (Kaushal and Singh, 2025).

**Governance Fit (R8)** integrates procedural transparency, documentation, and auditable compliance properties. *Justification:* The Guaranteed Safe AI framework argues that verifiable safety properties should be checkable over system logs (Dalrymple et al., 2024).

## 4. A Symbiotic Human-AI Framework

We propose a three-layer framework designed to satisfy R1–R8, with each layer specified in sufficient technical detail to enable implementation and evaluation. The layers form an iterative loop: information flows upward from observation to synthesis to ratification, while governance constraints flow downward.

### 4.1. Layer 1: Observation and Diversity Amplification

Layer 1 ingests parallel contributions and produces a structured map without generating consensus statements. The explicit aim is diversity amplification rather than compression, addressing concerns that optimization-driven systems prematurely narrow the space of considered options.

Contributions are embedded using a sentence transformer (e.g., all-MiniLM-L6-v2). We recommend hierarchical agglomerative clustering with Ward linkage as the default algorithm, since it produces a full dendrogram that allows exploration at multiple granularity levels without precommitting to a fixed number of clusters. The dendrogram can be cut at a level determined by the gap statistic (Tibshirani et al., 2001) or by silhouette analysis across a range of candidate cuts. K-means may be preferred when computational efficiency is paramount (e.g., at the 50,000-participant scale), with $k$ selected via the elbow method cross-validated against silhouette scores. In both cases, stability diagnostics across multiple random initializations should be reported. We employ consensus clustering to identify stable groupings and report semantic coherence measures (mean intra-cluster similarity) alongside cluster assignments. We compute distributional entropy over clusters:

$$H(C^t) = -\sum_{k=1}^{K} p_k \log p_k$$

where $p_k = |C^t_k| / \sum_j |C^t_j|$. Higher entropy indicates more balanced representation across clusters. This follows from the information-theoretic property that entropy is maximized when the distribution is uniform: if all clusters have equal membership, $H$ is at its maximum $\log K$, while a single dominant cluster drives $H$ toward zero. In our context, balanced cluster sizes indicate that no single viewpoint dominates the representation, which is a necessary (though not sufficient) condition for diversity. We also track minority cluster representation rate and identify bridge contributions that exhibit high cross-cluster similarity, signaling potential reframing opportunities (addressing R5). Where structured argumentation pipelines are available, Layer 1 can additionally extract argument components (claims, premises, rebuttals) to produce richer representations than flat topic clusters alone (Chen et al., 2024; Khader et al., 2024).

This design operationalizes Page's simultaneity argument by transforming parallel input into an attention guide for humans (Page, 2025). Rather than requiring sequential processing, the system provides a structured overview that helps facilitators allocate attention strategically. The approach also aligns with network science insights on preserving informational diversity (Centola, 2022): by making diversity visible rather than hiding it behind aggregation, the system creates conditions for exploration of alternatives.

### 4.2. Layer 2: Facilitation with Provenance

Layer 2 generates candidate synthesis statements subject to explicit constraints and full provenance tracking. Each candidate must include representation of major clusters and at least one minority critique, a “tension” line naming unresolved value conflicts rather than erasing them, and provenance linking phrases to supporting inputs.

#### 4.2.1. Formal Metric Definitions

We provide rigorous definitions for the multi-objective components, addressing concerns about binary indicators that may overcount weak matches by incorporating graded scoring.

**Coverage** measures graded cluster recall, aggregating top-$k$ clause-to-cluster similarities rather than relying solely on a binary threshold:

$$\text{Coverage}(s) = \sum_{k=1}^{K} w_k \max_{c \in s} \text{sim}(c, C_k)$$

where $w_k$ reflects cluster importance (uniform for equal treatment, or proportional to cluster size) and $\text{sim}(c, C_k)$ is the maximum cosine similarity between clause embedding $c$ and any contribution embedding in cluster $C_k$. This graded formulation rewards stronger representation rather than merely crossing a binary threshold. For deployment, we recommend flagging clusters where $\max_{c \in s} \text{sim}(c, C_k) < 0.7$ as under-represented, based on typical sentence-transformer similarity distributions, while noting that this threshold should be calibrated to human judgments of adequate representation through pilot annotation studies.

**Diversity** uses salience-aware minority weighting rather than relying solely on cluster size be-low the median. We define minority clusters as those below median size *or* those flagged as underrepresented by participants or facilitators, and weight each by a salience score combining size rarity, participant-flagged importance, and topical distinctiveness:

$$\text{Diversity}(s) = \frac{\sum_{k \in \mathcal{M}} \text{sal}(k) \cdot \max_{c \in s} \text{sim}(c, C_k)}{\sum_{k \in \mathcal{M}} \text{sal}(k)}$$

where $\mathcal{M}$ is the set of minority clusters and $\text{sal}(k)$ is the salience weight for cluster $k$.

**Erasure** penalizes omission of high-salience critiques:

$$\text{Erasure}(s) = \sum_{x_i \in X_{\text{sal}}} \text{sal}(x_i) \cdot (1 - \max_{c \in s} \text{sim}(c, x_i))$$

where $X_{\text{sal}}$ denotes contributions flagged as salient critiques. The continuous formulation $(1 - \text{sim})$ provides a graded penalty proportional to the degree of under-representation.

The selection objective combines these metrics:

$$\max_{s \in S'} \alpha \cdot \text{Endorse}(s) + \beta \cdot \text{Cov}(s) + \gamma \cdot \text{Div}(s) - \delta \cdot \text{Era}(s) \tag{1}$$

#### 4.2.2. Preference-Conditioned Trade-off Control

The weights $(\alpha, \beta, \gamma, \delta)$ encode normative assumptions that must be governed through deliberate processes. Rather than fixing these weights *a priori*, we propose leveraging preference-conditioned alignment mechanisms. Following PARM (Lin et al., 2025), a unified reward model is conditioned on preference vectors $\mathbf{p} = (\alpha, \beta, \gamma, \delta)$ normalized to sum to 1. The PBLoRA mechanism modulates model parameters:

$$\mathbf{W}_{\text{adapted}} = \mathbf{W}_0 + \mathbf{B} \cdot f(\mathbf{p}) \cdot \mathbf{A}$$

where $\mathbf{B}$ and $\mathbf{A}$ are low-rank matrices and $f(\mathbf{p})$ maps the preference vector to scaling coefficients. This enables facilitators to adjust trade-offs dynamically based on deliberation context (e.g., weighting diversity more heavily in early exploratory rounds, endorsement more in final convergence rounds) and exposes these controls for governance oversight.

Weight selection should itself be a participatory process. We propose that facilitators and participant representatives jointly define acceptable weight ranges before deliberation begins, documented as a signed "deliberation constitution" specifying bounds on each weight and conditions under which adjustments are permitted.

#### 4.2.3. Clause-Level Provenance Pipeline

We address the concern that semantic similarity may not reflect causal contribution through a five-stage pipeline: (i) *clause segmentation* via dependency parsing targeting 10–25 tokens per clause; (ii) *attribution scoring* using a cross-encoder (e.g., fine-tuned DeBERTa-v3) for pairwise clause-to-contribution similarity; (iii) *causal diagnostics* via knockout regeneration, re-generating clauses after removing candidate sources to measure marginal influence; (iv) *explicit citation extraction* when participants directly reference others; and (v) *confidence scoring* based on agreement between similarity and causal evidence, calibrated against human-annotated gold standards.

Each clause $c$ stores a support set $I(c)$ with confidence-weighted attribution scores and uncertainty bounds, enabling forward queries ("what did my contribution influence?") and backward queries ("why does this clause appear?") with appropriate epistemic humility (e.g., "This clause is likely influenced by contributions 47 and 92, confidence 0.78 ±0.09").

### 4.3. Layer 3: Human Primacy and Resolution

Layer 3 enforces human primacy through explicit governance controls, addressing R3 and R6.

#### 4.3.1. Contestability Workflow

We specify concrete contestability mechanisms with governance parameters.

**Challenge submission.** Participants can flag any clause as misrepresenting their contribution, with a baseline rate limit of 3 challenges per participant per round to prevent flooding. Rate limits are *equity-aware*: participants from underrepresented clusters receive higher challenge allowances (e.g., $3 + \lfloor 2 \cdot (1 - p_k) \rfloor$ where $p_k$ is their cluster's representation proportion). Each challenge includes the flagged clause, the participant's original contribution, and a brief explanation (maximum 100 words).

**Alternative framing.** Participants may request alternative phrasings for contested clauses. The system generates 2–3 alternatives using constrained decoding that maintains provenance links while varying surface form.

**Veto rights.** Facilitators hold veto authority over any output with documented rationale. Participant-level veto requires threshold support (e.g., 20% of active participants) to trigger reconsideration.

**Dispute resolution.** Target turnaround is 4 hours for challenge review, with an escalation path to human facilitators for unresolved disputes. All challenge-resolution sequences are logged for governance audit.

**Worked Example.** Consider a deliberation on urban transport policy with 200 participants. Layer 1 identifies five theme clusters, including a minority cluster (12 participants) advocating for disability access with salience weight $\text{sal}(k) = 0.85$ (high topical distinctiveness). Layer 2 generates a candidate statement; the provenance display shows that the disability-access cluster has low graded coverage ($\max$ sim $= 0.52$). A participant from this cluster submits a challenge: "Clause 3 mentions 'inclusive design' but does not address wheelchair-accessible vehicle requirements, which was the core of our argument." The system generates three alternative phrasings for Clause 3. Knockout regeneration confirms causal influence (score 0.72). The participant selects an alternative, the updated statement is re-scored on the multi-objective function (Eq. 1), and the modification is logged with full provenance for audit. This cycle completes within one round without facilitator intervention, though facilitator review remains available.

### 4.3.2. Ratification Process

Final endorsement requires explicit human decision with a recorded procedural log. We recommend multi-stage ratification: initial review by a randomly-selected participant subset, revision based on feedback, and full-group endorsement vote with a minimum participation threshold. This layer directly addresses critiques that automated mediation risks replacing political agency with optimized text (Volpe, 2025; Oleart and Palomo Hernández, 2025), and mirrors the design logic of Human-AI Deliberation systems where disagreement triggers structured iteration rather than forced acceptance (Ma et al., 2025).

## 4.4. Robustness and Security

We specify concrete robustness tests, informed by recent work on adversarially robust text attribution (Sai Teja et al., 2025).

**Adversarial Attribution Tests.** We inject near-duplicate or paraphrase attacks into contribution pools and measure provenance drift. The acceptable drift threshold is $<$10% change in top-3 attributed contributions under semantically equivalent perturbations. Drawing on insights from adversarial authorship segmentation research (Sai Teja et al., 2025), we additionally test for invisible-character attacks and stylometric mimicry attempts.

**Prompt Injection Detection.** Layer 1 employs classifier-based detection of injection attempts. Detected attempts are quarantined with logged review and preserved appeal rights.

| Dimension | Human | LLM-only | Symbiotic |
|---|---|---|---|
| Scale (R1) | Low–med | Med–high | High |
| Diversity (R2) | Manual | Implicit | Explicit |
| Contestability (R3) | High | Low | High |
| Provenance (R4) | Memory | Absent | Required |
| Robustness (R7) | Social norms | Minimal | Tested |
| Legitimacy basis | Process | Fragile | Strengthened |

Table 1: Comparison of deliberation support approaches. "Explicit," "Required," and "Tested" indicate formally specified, enforced, and adversarially evaluated properties.

**Flood Detection.** Burstiness detection flags rapid submission patterns exceeding $3\sigma$ from participant means. Lexical diversity monitoring identifies suspiciously homogeneous contribution batches.

**Manipulation Gaming.** To address provenance gaming, we employ contribution timing weighting, cross-round consistency checking, and facilitator override capabilities. The knockout regeneration tests in Section 4.2.3 provide additional resistance, since causal influence is harder to fake than surface similarity.

Crucially, the existence of filtering must be transparent: participants should know that some content may be flagged, understand the criteria, and have access to appeals processes.

## 4.5. Formal Safety Properties

Drawing on the Guaranteed Safe AI framework (Dalrymple et al., 2024), we define verifiable safety properties checkable over system logs: minimum minority-cluster coverage (every cluster $k \in \mathcal{M}$ achieves $\max$ sim $> \tau_{\min}$), contestability SLA (all challenges receive response within the specified turnaround), provenance completeness (every clause has $\geq 1$ attributed source with confidence $> 0.5$), and erasure bound (total erasure score below a governance-defined threshold). These properties can be verified automatically over deliberation logs and reported in post-round audit summaries.

## 4.6. Framework Comparison

Table 1 compares the three approaches. Human-only deliberation preserves contestability but faces hard scalability limits. LLM-only mediation achieves scale but lacks provenance and leaves legitimacy fragile. The symbiotic framework restores accountability properties while retaining scale, at the cost of added complexity whose net benefit the proposed ablation design (Section 6) is intended to test.

## 5. System Architecture and Language Resources

The architecture separates concerns across the three layers. An input ingestion module accepts contributions in parallel and stores them with timestamps, participant identifiers, and metadata. Layer 1 components produce theme maps and diversity dashboards through embedding, clustering, and visualization. Layer 2 components generate candidate statements with clause-level provenance through prompted language models with structured output requirements. Layer 3 components provide review interfaces, voting mechanisms, and audit logging.

This separation enables modular development and testing. For multilingual deliberations, Layer 1 requires cross-lingual embeddings (e.g., multilingual sentence transformers), and Layer 2 requires generation models with multilingual capability and translation quality monitoring. Accessibility provisions should include ASR integration for voice contributions and screen-reader-compatible provenance displays. Such multilingual and cross-dialectal capability is essential for deploying deliberation systems in linguistically diverse contexts, where stance and sentiment expression patterns vary across dialects and cultural communities (Laabar and Zaghouani, 2024; Al Heraki and Zaghouani, 2025).

### 5.1. Deliberation Log Schema

Each contribution record includes: anonymized participant identifier and role metadata, timestamp and round identifier, raw text content, optional stance label and claim type, evidence pointers linking to external sources, emotion tags capturing affective dimensions, and safety markers flagging suspected spam or coordinated flooding. Output artifacts include theme map snapshots, candidate statements with provenance graphs, human review logs, and final endorsed statements with complete provenance chains. Privacy-preserving logging should employ differential privacy mechanisms for aggregate statistics and tiered consent models specifying visibility of provenance information.

This schema transforms deliberation from ephemeral discussion into a reusable resource for research on summarization, argument structure, stance dynamics, and democratic governance.

### 5.2. Provenance Representation

For each clause $c$ in statement $s$, the system stores a support set $I(c)$ of contributing input identifiers with weights combining semantic similarity scores, causal influence estimates from knockout tests, explicit citation counts, and facilitator annotations.

The provenance graph is versioned across rounds, allowing temporal analysis of how representations evolve. This representation enables the forward and backward queries described in Section 4.2.3, which are essential for contestability.

## 6. Evaluation Protocol

We propose an evaluation protocol informed by findings on LLM judge limitations and deliberation benchmark results, designed for implementation in future empirical studies.

### 6.1. Study Design

A three-arm experimental design compares: (a) human facilitation baseline using trained moderators, (b) LLM mediation without provenance or veto (replicating the Habermas Machine approach), and (c) the symbiotic framework with all three layers active.

**Pre-registration.** We recommend pre-registration of hypotheses, primary and secondary metrics, and analysis plans to guard against post-hoc rationalization.

**Power Analysis.** For the primary outcome of endorsement difference, assuming medium effect size ($d = 0.5$) and $\alpha = 0.05$, a two-tailed test requires $n = 64$ per condition for 80% power. For diversity retention ($d = 0.4$), $n = 100$ per condition. We recommend $n = 150$ per condition to enable subgroup analyses.

**Blinding and Manipulation Checks.** Where feasible, participants should be blinded to mediation source to mitigate AI halo effects (Tessler et al., 2024). Post-deliberation debriefing should include manipulation checks for blinding effectiveness and measure perceptions of legitimacy and traceability separately from content quality, using validated psychometric scales for procedural justice and institutional trust.

**Ablation Design.** To quantify each layer's contribution, we recommend additional conditions: symbiotic framework without provenance display, and symbiotic framework without contestability mechanisms. These ablations enable causal attribution of legitimacy and satisfaction improvements to specific framework components rather than the bundle as a whole.

### 6.2. Metrics

**Outcome Metrics.** Endorsement measures participant preference through Likert ratings and pairwise comparisons. Quality assessment uses external evaluator ratings on clarity, informativeness, and fairness with standardized rubrics. Legitimacy is assessed using validated procedural justice scales, stratified by demographics and political identity.

**Process Metrics.** Diversity retention tracks cluster coverage and minority inclusion across rounds. Polarization shift compares pre- and post-deliberation stance distributions. Contestability use records challenge frequency, resolution rates, time-to-resolution, and equity of challenge uptake. Provenance quality involves human audit of attribution precision and coverage at the clause level.

**Human Raters over LLM Judges.** Based on CounselBench findings (Li et al., 2025), we recommend against using LLM judges for nuanced criteria such as representational fairness. Human raters with span-level audit protocols should assess fairness of representation and provenance accuracy, with inter-rater reliability reporting (Krippendorff's $\alpha > 0.7$ threshold).

**Adversarial Stress Tests.** Live red-teaming sessions with prompt injection, flooding, and paraphrase pollution should accompany controlled evaluations to stress-test attribution stability.

## 6.3. Participant Recruitment and Testing Costs

A critical practical consideration for empirical validation is the cost and logistics of participant recruitment. We recommend recruiting participants through established crowdsourcing platforms such as Prolific, which supports demographic prescreening and has been widely used in deliberation research. For the recommended $n = 150$ per condition across three conditions plus two ablation arms (five conditions total), the study requires approximately 750 participants. At current Prolific rates of approximately $12–$15 per hour for a 60-minute deliberation session, participant compensation alone would cost between $9,000 and $11,250. API costs for LLM inference, embedding, and provenance computation add approximately $200–$400 for a 750-participant study at current pricing (see Table 2). Facilitator training and compensation for the human-mediation baseline condition would add approximately $2,000–$3,000. The total estimated cost for a full five-condition study is therefore in the range of $11,000–$15,000, which is feasible for a standard research grant. For larger-scale pilots (e.g., 5,000 participants for external validity), costs scale roughly linearly with participant count but sublinearly for API and facilitator overhead.

## 6.4. Cost and Latency Analysis

Table 2 provides estimates across deployment scales based on component costs.

The addition of knockout regeneration tests increases costs by approximately 50–60% over similarity-only provenance but provides substantially stronger attribution evidence. Facilitator workload scales sublinearly due to automated filtering but remains the practical ceiling for rapid-turnaround deployments.

| Component | 500 | 5,000 | 50,000 |
|---|---|---|---|
| Layer 1 (embed/cluster) | 2 min | 15 min | 2.5 hr |
| Layer 2 (generation) | 5 min | 20 min | 1.5 hr |
| Layer 2 (provenance) | 10 min | 1.5 hr | 12 hr |
| Layer 2 (knockout tests) | 15 min | 2 hr | 18 hr |
| Facilitator review | 30 min | 4 hr | 24+ hr |
| Est. API cost | $8 | $80 | $800 |

Table 2: Estimated processing times and costs by participant scale. Knockout tests add latency but can be parallelized. At 50,000 participants, batch processing with incremental updates is recommended.

## 6.5. Reproducibility

To support cumulative science, we recommend releasing anonymized deliberation logs following the schema described above, statement candidates and provenance graphs for all conditions, evaluation rubrics with detailed coding instructions, human-annotated clause-source gold standards for provenance calibration, and analysis code for computing all metrics.

# 7. Discussion

The framework is designed as infrastructure for empirical testing rather than a final system. The deliberation log schema, formal metrics, verifiable safety properties, and evaluation protocol provide the scaffolding for controlled studies. The ablation design (Section 6) is specifically intended to isolate whether provenance and contestability machinery delivers net benefits over simpler approaches. Three research questions strike us as most pressing. First, do equity-aware contestability mechanisms achieve comparable uptake across demographic groups, or do structural barriers persist despite algorithmic accommodation? Second, does explicit provenance increase or erode participant trust, and under what conditions does attribution transparency become counterproductive? Third, how should metric weights be calibrated across deliberative contexts, and can participatory weight-selection processes themselves satisfy deliberative norms?

We note that the framework's formal specification of diversity, provenance, and contestability requirements is complementary to existing empirical work on stance detection, polarization analysis, and discourse annotation in multilingual social media contexts (Laabar and Zaghouani, 2024; Al Heraki and Zaghouani, 2025). The deliberation log schema we propose could serve as a bridge between computational approaches to opinion mining and the normative requirements of democratic deliberation, enabling researchers to study how diverse perspectives are represented and transformed during AI-mediated synthesis.

## 8. Limitations

This paper presents a framework rather than implementation results. We view this as a necessary intermediate step: existing systems either lack formal specification of fairness-relevant properties or conflate endorsement with legitimacy. Our contribution is a specification precise enough to implement, critique, and empirically test. Key open challenges include provenance gaming by determined adversaries beyond what knockout tests currently address; the need for annotation studies to calibrate metric thresholds against human judgments of fair representation; empirical validation that salience-aware minority definitions better capture normative fairness than size-based criteria; and field deployment challenges including participant dropout, multilingual quality assurance, and institutional integration. We recommend that deployments establish institutional memoranda of understanding specifying data handling, audit access, and independent review provisions.

## 9. Conclusion

Scalable AI-mediated deliberation is feasible only when contestability and governance are built into the technical core rather than treated as afterthoughts. We have proposed a symbiotic framework that specifies formal metrics, causal provenance, preference-conditioned control, equity-aware contestability, adversarial robustness protocols, and an ablation-ready evaluation design. The framework keeps AI in a facilitative role and preserves human primacy over interpretation and endorsement. We invite the community to use this specification as a basis for implementation, controlled evaluation, and the creation of deliberative NLP resources.

## 10. Ethical Considerations

This framework addresses AI-augmented democratic deliberation, where ethical design is intrinsic to the technical contribution. Systems that mediate deliberation inevitably shape whose voices are amplified and how consensus is constructed. Although contestability and provenance mechanisms are intended to render this influence visible and challengeable, transparency can become performative if institutions deploy interface features without meaningful responsiveness to participant feedback (Oleart and Palomo Hernández, 2025). Deployments should therefore include independent auditing with access to governance logs, publicly reported compliance with defined safety properties, participant exit surveys assessing whether contestation was substantively addressed, and binding contractual terms specifying how outputs may be used in downstream decisions.

Digital deliberation environments also risk reproducing structural inequalities. Participants with greater time, digital literacy, or institutional familiarity may dominate both contribution and challenge mechanisms. Equity-aware rate limits and salience-weighted diversity metrics offer partial mitigation, but procedural inclusion cannot be reduced to algorithmic adjustment. Responsible deployment requires accessibility provisions, multilingual support, active recruitment of underrepresented communities, and disaggregated monitoring of participation patterns.

Clause-level provenance introduces a tension between accountability and privacy. Detailed attribution strengthens traceability but may discourage candid participation, particularly on sensitive issues. Tiered consent models, differential privacy for aggregate statistics, and data minimization after deliberation are necessary safeguards.

Finally, the same mechanisms intended to enhance legitimacy could be misused. Diversity metrics may be gamed to simulate inclusion, provenance displays selectively curated, and clustering infrastructures repurposed for opinion surveillance. Deployment is inappropriate where participants lack genuine influence over outcomes. Future work must remain attentive to the distinction between enabling meaningful collective agency and extracting the appearance of consent.

## Acknowledgments

This work was made possible by the National Priorities Research Program (NPRP) grant NPRP14C-0916-210015 from the Qatar National Research Fund (QNRF), a member of the Qatar Research, Development and Innovation Council (QRDI).